\documentclass{article}

\usepackage[preprint]{neurips_2025}

\usepackage[utf8]{inputenc} % allow utf-8 input
\usepackage[T1]{fontenc}    % use 8-bit T1 fonts
\usepackage{hyperref}       % hyperlinks
\usepackage{url}            % simple URL typesetting
\usepackage{booktabs}       % professional-quality tables
\usepackage{amsfonts}       % blackboard math symbols
\usepackage{nicefrac}       % compact symbols for 1/2, etc.
\usepackage{microtype}      % microtypography
\usepackage{xcolor}         % colors
\usepackage{graphicx}
\usepackage{amsmath} 
\usepackage{multirow}
\usepackage{booktabs} 
\usepackage{multicol}
\usepackage{xspace}

\def\eg{\textit{e.g.}\xspace}

\title{FreeLoRA: Enabling Training-Free LoRA Fusion for Autoregressive Multi-Subject Personalization}

\author{
Peng Zheng$^{1,2}$ \quad
Ye Wang$^1$ \quad
Rui Ma$^{1,*}$ \quad
Zuxuan Wu$^{2,3,*}$ \\
$^1$School of Artificial Intelligence, Jilin University \\
$^2$Shanghai Innovation Institute \\
$^3$School of Computer Science, Fudan University \\
{\tt\small zhengpeng22@mails.jlu.edu.cn}
}
\begin{document}

\maketitle

\begin{figure}[ht]
    \centering
    \includegraphics[width=\linewidth]{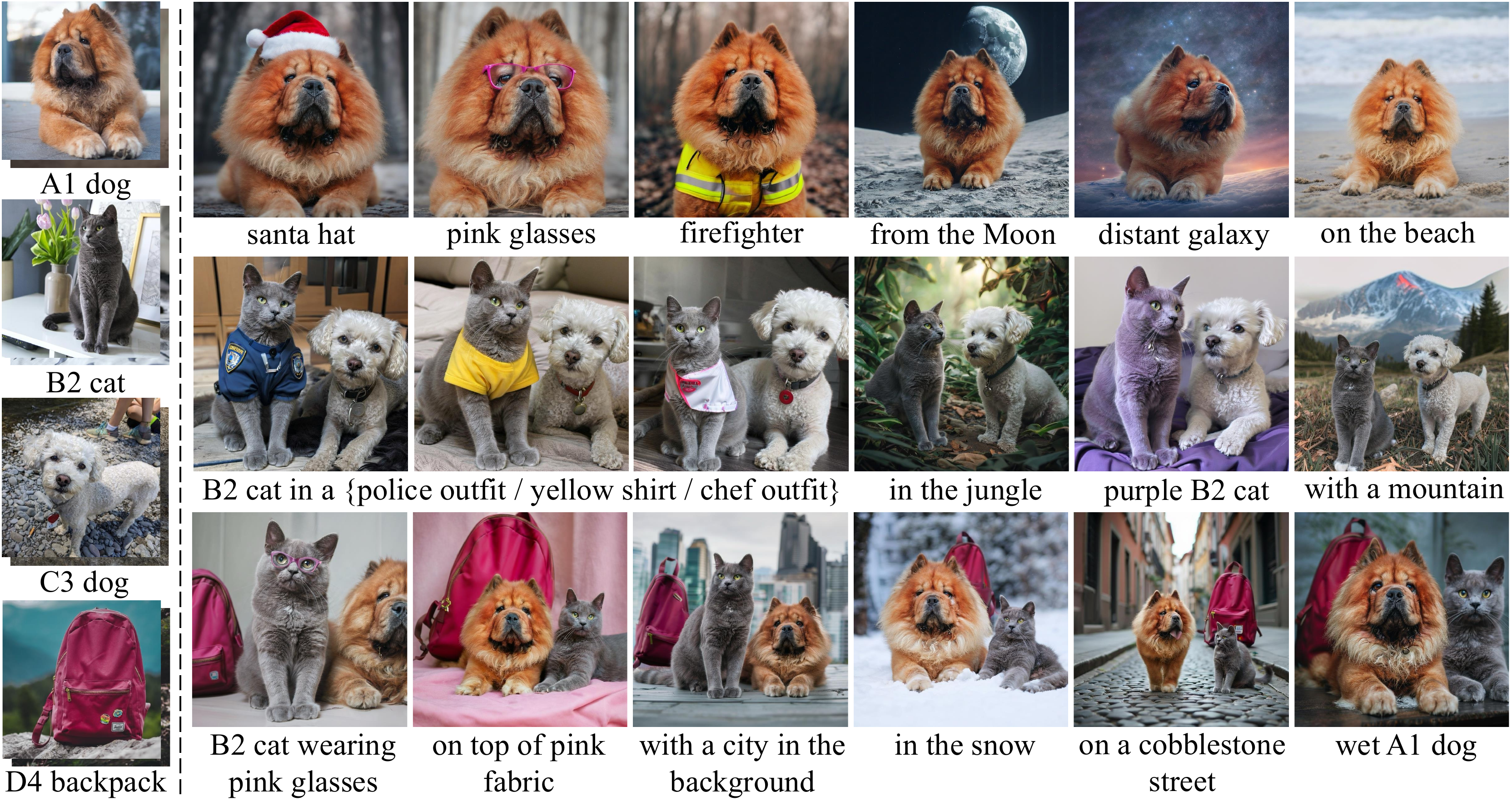}
    \caption{Personalized image generation with FreeLoRA. Each LoRA module is independently tuned on a reference subject (left). At inference time, FreeLoRA fuses them without additional re-tuning, enabling both single-subject and multi-subject personalization across diverse prompts.}
    \label{fig:teaser}
\end{figure}

\begin{abstract}
Subject-driven image generation plays a crucial role in applications such as virtual try-on and poster design. Existing approaches typically fine-tune pretrained generative models or apply LoRA-based adaptations for individual subjects. However, these methods struggle with multi-subject personalization, as combining independently adapted modules often requires complex re-tuning or joint optimization. We present FreeLoRA, a simple and generalizable framework that enables training-free fusion of subject-specific LoRA modules for multi-subject personalization. Each LoRA module is adapted on a few images of a specific subject using a Full Token Tuning strategy, where it is applied across all tokens in the prompt to encourage weakly supervised token-content alignment. At inference, we adopt Subject-Aware Inference, activating each module only on its corresponding subject tokens. This enables training-free fusion of multiple personalized subjects within a single image, while mitigating overfitting and mutual interference between subjects.
Extensive experiments show that FreeLoRA achieves strong performance in both subject fidelity and prompt consistency.
\end{abstract}
\section{Introduction}
Subject-driven image generation~\cite{ruiz2023dreambooth,gal2022image,li2023blip}, which aims to synthesize images of specific entities based on a few reference images, has become increasingly important in applications such as virtual try-on, personalized content creation, and advertising design. This task requires models to preserve the visual identity of a subject, such as a pet or object, while flexibly adapting to diverse prompts.

A typical solution is to fine-tune large-scale pretrained generative models on subject-specific data~\cite{ruiz2023dreambooth,Hao2023ViCo,chung2025fine,kumari2023multi}. More recently, low-rank adaptation (LoRA)~\cite{hu2022lora} has emerged as a parameter-efficient alternative, enabling lightweight customization without full fine-tuning~\cite{wu2025proxy,zhou2024magictailor}. Each subject is typically represented by an independently trained LoRA module. However, existing approaches struggle with multi-subject personalization: merging multiple subject-specific modules often requires additional re-tuning, which increases computational cost and may degrade generation quality~\cite{gu2023mix,yang2024lora,huang2025resolving,liu2023cones,meral2024clora,kumari2023multi}.

Meanwhile, another line of research investigates training-based pipelines that learn subject-driven generation from large-scale curated datasets~\cite{wu2025less,mou2025dreamo,xiao2024omnigen,tan2024ominicontrol}. While these methods support multi-subject composition, they typically rely on a complex data pipeline and costly training, and may compromise subject fidelity due to limited coverage or insufficient variation.

In this work, we seek a simpler and more efficient solution: a framework that enables training-free fusion of subject-specific LoRA modules at inference time. We introduce FreeLoRA, a lightweight approach for multi-subject personalization within visual autoregressive models. Each LoRA module is adapted on a small set of reference images and inserted into the key and value (KV) projections of the cross-attention layers.
During tuning, we adopt a Full Token Tuning strategy, applying each LoRA module to the KV projections of all text tokens in the prompt. This allows the model to learn token-content correspondence in a weakly supervised manner.
During inference, we apply Subject-Aware Inference by activating each LoRA module only on the KV projections associated with its corresponding subject token. This enables multi-subject fusion without additional re-tuning, while mitigating overfitting and avoiding interference across subjects. The results are shown in Figure~\ref{fig:teaser}.

We validate FreeLoRA across diverse subject compositions and demonstrate strong performance in subject fidelity, prompt consistency, and inference-time flexibility. Our approach is implemented on Infinity-2B~\cite{han2024infinity}, a large-scale visual autoregressive (AR) model~\cite{li2025autoregressive,weber2024maskbit,tian2025visual,yu2024randomized}, which holds great promise for enabling multimodal large language models (MLLMs)~\cite{team2023gemini,zhou2024transfusion,shi2024llamafusion,tong2024metamorph,wu2024vila,xie2024show,wang2024emu3,team2024chameleon}. Notably, the FreeLoRA framework remains compatible with various Transformer-based generative models, requiring only model-specific LoRA modules.

In summary, the contributions of this study are as follows:
\begin{itemize}
    \item We propose FreeLoRA, a novel framework that enables training-free fusion of subject-specific LoRA modules through Subject-Aware Inference (SAI). This approach facilitates multi-subject personalization without additional re-tuning, allowing for flexible and efficient integration of multiple subjects during inference. 
    \item We introduce Full Token Tuning, a training strategy that applies LoRA modules across all tokens. This design allows the model to automatically learn the alignment between textual tokens and visual content. Combined with SAI, it enables accurate extraction of subject-specific features while mitigating overfitting to the reference images.
    \item To the best of our knowledge, our work is the first to achieve multi-subject personalization by fine-tuning a visual autoregressive model, demonstrating the potential of tuning-based approaches in visual autoregressive frameworks.
\end{itemize}
\section{Related Work}

\subsection{Tuning-Based Subject Personalization}

\textbf{Single-subject personalization.}
A common approach for subject personalization is to fine-tune a pretrained model on a few images of a target entity.
DreamBooth~\cite{ruiz2023dreambooth} adapts diffusion models to preserve subject identity while maintaining prompt flexibility.
Textual Inversion~\cite{gal2022image} learns subject-specific word embeddings without modifying the model backbone.
Community efforts~\cite{lora_stable} have leveraged LoRA~\cite{hu2022lora} for efficient image generator personalization.
BLIP-Diffusion~\cite{li2023blip} proposes a two-stage pipeline involving multimodal representation learning and subject representation learning to facilitate efficient tuning.
While most prior works~\cite{shi2024instantbooth} use diffusion models~\cite{rombach2022high, podell2023sdxl, dhariwal2021diffusion} for personalization, recent advances in visual autoregressive (AR) models~\cite{han2024infinity, li2025autoregressive, weber2024maskbit, tian2025visual, yu2024randomized} have sparked growing interest in AR-based subject personalization.
Methods such as Proxy-Tuning~\cite{wu2025proxy}, PersonalAR~\cite{sun2025personalized}, and ARBooth~\cite{chung2025fine} explore subject-driven generation within AR models, demonstrating strong identity fidelity and prompt consistency.
However, these AR-based methods have not yet explored multi-subject personalization.

% \textbf{Single-subject tuning.}
% A common approach for Subject Personalization involves fine-tuning a pretrained model on a few images of a target entity. 
% DreamBooth~\cite{ruiz2023dreambooth} adapts diffusion models to preserve subject identity while maintaining prompt flexibility. Textual Inversion~\cite{gal2022image} learns subject-specific word embeddings, avoiding updates to the model backbone. Community efforts~\cite{lora_stable} have applied LoRA~\cite{hu2022lora} to image generators for efficient personalization. VICO~\cite{Hao2023ViCo} introduces an image attention module to enable plug-and-play subject control. BLIP-Diffusion~\cite{li2023blip} proposes a two-stage pipeline involving multimodal representation learning and subject representation learning to facilitate efficient fine-tuning. While most prior works~\cite{shi2024instantbooth} utilize diffusion models~\cite{rombach2022high, podell2023sdxl, dhariwal2021diffusion} for personalization, recent advances in visual autoregressive (AR) models~\cite{han2024infinity, li2025autoregressive, weber2024maskbit, tian2025visual, yu2024randomized} have sparked interest in AR-based subject tuning. Methods such as Proxy-Tuning~\cite{wu2025proxy}, PersonalAR~\cite{sun2025personalized}, and ARBooth~\cite{chung2025fine} explore subject-driven generation within AR models, demonstrating strong identity fidelity and prompt consistency. However, these AR-based approaches have yet to explore multi-subject personalization.

\textbf{Multi-subject personalization.}
Custom Diffusion~\cite{kumari2023multi} extends personalization to multi-subject settings by merging multiple fine-tuned models through closed-form constrained optimization.
Mix-of-Show~\cite{gu2023mix} introduces gradient fusion and regionally controllable sampling to combine multiple LoRA modules.
CLoRA~\cite{meral2024clora} adopts contrastive learning for LoRA composition, while CIDM~\cite{dong2024continually} proposes concept-incremental learning for continual subject integration.
Although these tuning-based methods~\cite{kumari2023multi, gu2023mix, meral2024clora, dong2024continually, yang2024lora, liu2023cones} successfully support multi-subject personalization, they typically require additional optimization to merge LoRA modules or fine-tuned models, which may hinder practical deployment and degrade image quality.
Some recent efforts~\cite{zhong2024multi, shah2024ziplora, ouyang2025k} explore training-free fusion of multiple LoRAs, though these are not explicitly designed for multi-subject generation.

\subsection{Training-Free Subject Personalization}

Training-free methods~\cite{ding2024freecustom, shin2024large, zhang2025bringing, feng2025personalize, kang2025flux, nam2024dreammatcher} aim to personalize generative models without any training.
FreeCustom~\cite{ding2024freecustom} introduces a multi-reference self-attention mechanism that injects reference features directly into the self-attention layers, enabling training-free multi-subject generation.
Diptych Prompting~\cite{shin2024large} shows that inpainting-based text-to-image models can perform zero-shot subject-driven generation and proposes a reference attention enhancement mechanism.
Personalize Anything~\cite{feng2025personalize} achieves subject control using time-step adaptive token replacement and patch perturbation strategies, supporting layout-guided and multi-subject generation.
While these approaches offer efficient inference, training-free paradigms often struggle to maintain subject fidelity and prompt flexibility due to the constraints of the underlying generative model.

\subsection{Training-Based Subject Personalization}

Another line of work~\cite{he2025anystory} focuses on training subject-driven generative models. ELITE~\cite{wei2023elite} learns a mapping network that projects images into the text embedding space, enabling fast personalized generation without per-subject tuning. Subject-Diffusion~\cite{ma2024subject} supports open-domain subject-driven generation by incorporating coarse localization and fine-grained image control. MIP-Adapter~\cite{huang2025resolving} and MS-Diffusion~\cite{wang2024ms} focus on training the cross-attention layers of pretrained generative models for subject personalization. Broader frameworks like OminiControl~\cite{tan2024ominicontrol}, OmniGen~\cite{xiao2024omnigen}, DreamO~\cite{mou2025dreamo}, and UNO~\cite{wu2025less} aim to unify image customization through large-scale training. While these methods demonstrate powerful personalization capabilities, they often demand substantial computational resources and may suffer from performance degradation when handling rare subjects.
\section{Preliminary: Visual Autoregressive Models}

Visual Autoregressive (AR) models have emerged as a powerful framework for image generation, where an image $\mathbf{I} \in \mathbb{R}^{H \times W \times 3}$ is encoded into a sequence of discrete tokens $\mathbf{z} = \{z_1, z_2, \dots, z_N\}$ using a visual tokenizer. Conditioned on a textual prompt $\mathbf{c}$, the model learns the joint distribution:
\begin{equation}
P(\mathbf{z} \mid \mathbf{c}) = \prod_{i=1}^{N} P(z_i \mid z_{<i}, \mathbf{c}),
\end{equation}
and is trained to minimize the negative log-likelihood:
\begin{equation}
\mathcal{L}_{\text{AR}} = -\sum_{i=1}^{N} \log P(z_i \mid z_{<i}, \mathbf{c}).
\end{equation}

We adopt Infinity-2B~\cite{han2024infinity} as our backbone, which improves upon conventional AR models via a hierarchical bitwise residual quantizer. It predicts image tokens from coarse to fine scales, enabling high-resolution synthesis. While we utilize this framework for implementation, our method is model-agnostic and applicable to any transformer-based AR image generator.

% \subsection{Cross-Attention and LoRA Injection Points}

% In AR image generation, the cross-attention mechanism is used to align image tokens (queries) with prompt tokens (keys/values). For a given cross-attention layer, the attention is computed as:
% \[
% \text{Attention}(Q, K, V) = \text{softmax}\left(\frac{QK^\top}{\sqrt{d}}\right)V,
% \]
% where $Q \in \mathbb{R}^{L_q \times d}$ are query vectors from image tokens, and $K, V \in \mathbb{R}^{L_p \times d}$ are key and value vectors from text tokens.

% In our method, we inject Low-Rank Adaptation (LoRA)~\cite{hu2022lora} modules into the key and value projection matrices of the cross-attention layers. Specifically, for a weight matrix $W \in \mathbb{R}^{d \times d}$ in the key or value path, we replace it with:
% \[
% W' = W + \Delta W, \quad \Delta W = \alpha \cdot B A,
% \]
% where $A \in \mathbb{R}^{r \times d}$ and $B \in \mathbb{R}^{d \times r}$ are trainable low-rank matrices, and $\alpha$ is a scaling factor.

% We choose to insert LoRA only into the $K$ and $V$ projections—rather than $Q$ or output projections—to restrict each subject-specific module to influence only its associated prompt tokens. This selective and modular insertion forms the basis of our Subject-Aware Inference strategy and allows multiple LoRA modules to be fused at inference without conflict.
\section{The Proposed Method: FreeLoRA}

\begin{figure}
    \centering
    \includegraphics[width=\linewidth]{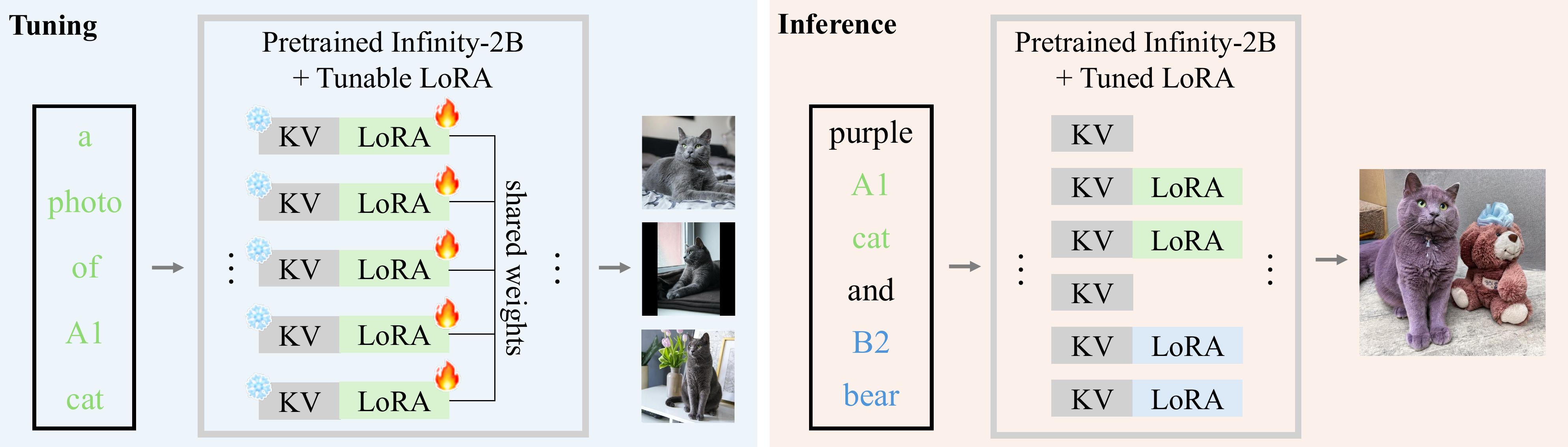}
    \caption{Overview of FreeLoRA. During tuning, we insert subject-specific LoRA modules into the key and value (KV) projections of cross-attention layers in a pretrained visual autoregressive (VAR) model Infinity-2B~\cite{han2024infinity}, while freezing the original parameters. We apply each LoRA module across all prompt tokens (Full Token Tuning) using a prompt-image pair (\eg, ``a photo of A1 cat''). During inference, each subject token in the prompt (\eg, ``A1 cat'' and ``B2 bear'' in ``purple A1 cat and B2 bear'') activates only its corresponding LoRA module (Subject-Aware Inference). This enables training-free fusion of multiple subjects in a single image.}
    \label{fig:pipeline}
\end{figure}

We present FreeLoRA, a simple and generalizable framework that enables training-free fusion of subject-specific LoRA modules for multi-subject personalization. FreeLoRA is tuned on a pretrained autoregressive model such as Infinity-2B~\cite{han2024infinity}, but remains compatible with other Transformer-based architectures. It involves tuning subject-specific LoRA modules and selectively activating them during inference based on the presence of corresponding subject tokens in the prompt. An overview of the pipeline is shown in Figure~\ref{fig:pipeline}.

\subsection{Cross-Attention with LoRA Injection Points}

In Transformer-based AR models, cross-attention layers enable the model to condition image token generation on a text prompt. Each layer computes:
\begin{equation}
\mathrm{Attention}(Q, K, V) = \mathrm{softmax}\left(\frac{Q K^\top}{\sqrt{d}}\right) V,
\end{equation}

where $Q \in \mathbb{R}^{L_q \times d}$ are queries from image tokens, and $K, V \in \mathbb{R}^{L_p \times d}$ are keys and values from prompt tokens.

To enable efficient and modular subject-specific adaptation, we introduce Low-Rank Adaptation (LoRA)~\cite{hu2022lora} into the key and value projections of cross-attention layers. Specifically, for a projection matrix $\mathbf{W} \in \mathbb{R}^{d \times d}$ (used for $K$ or $V$), the LoRA-augmented weight becomes:
\begin{equation}
\mathbf{W}' = \mathbf{W} + \Delta \mathbf{W}, \quad \Delta \mathbf{W} = \alpha \cdot \mathbf{B} \mathbf{A},
\end{equation}
where $\mathbf{A} \in \mathbb{R}^{r \times d}$ and $\mathbf{B} \in \mathbb{R}^{d \times r}$ are trainable low-rank matrices, with $r \ll d$, and $\alpha$ is a scaling factor. The base weight $\mathbf{W}$ is frozen.

By inserting LoRA only into $K$ and $V$, we isolate the influence of each subject-specific module to its corresponding prompt tokens. This design ensures modularity and paves the way for training-free composition by enabling selective activation at inference time.

\subsection{Full Token Tuning for Subject-Specific LoRA}

Given a set of reference images $\{\mathbf{I}_i\}_{i=1}^N$ of a single subject $S$, we pair each image with a prompt $\mathbf{T}_S$ that includes a unique identifier, such as ``a photo of A1 cat''. This forms our tuning dataset $\mathcal{D}_S = \{(\mathbf{I}_i, \mathbf{T}_S)\}_{i=1}^N$.

During tuning, we optimize only the LoRA parameters inserted into the key and value projections, while freezing all base model weights. The learning objective is the standard autoregressive negative log-likelihood:
\begin{equation}
\mathcal{L} = -\sum_{i=1}^N \log P_{\theta + \Delta \theta}(\mathbf{I}_i \mid \mathbf{T}_S),
\end{equation}
where $\theta$ denotes frozen pretrained parameters and $\Delta \theta$ denotes the LoRA parameters $(\mathbf{A}, \mathbf{B})$.

Crucially, we apply the subject LoRA module across \textit{all} prompt tokens, rather than restricting it to only the subject identifier. This \textbf{Full Token Tuning} (FTT) strategy allows the model to weakly associate different parts of the prompt with different visual content, such as ``cat'' with identity and ``photo'' with style or layout. Combined with Subject-Aware Inference (SAI) strategy introduced below, this approach improves alignment and generalization while mitigating overfitting to irrelevant components such as the background.

\subsection{Subject-Aware Inference for Multi-Subject Personalization}

At inference time, users may compose prompts that contain multiple subject tokens (\eg, ``a A1 cat and a B2 dog on a sunny street''). To support this, we propose \textbf{Subject-Aware Inference} (SAI): each LoRA module is activated only for the key and value projections corresponding to its associated subject tokens.

We extract a subject token set $\mathcal{S} = \{\text{A1 cat}, \text{B2 dog}, \dots\}$ from the prompt. For each subject $s \in \mathcal{S}$, we retrieve the trained LoRA parameters $(\mathbf{A}_s, \mathbf{B}_s)$ and compute:
\begin{equation}
\mathbf{W}'_s = \mathbf{W} + \alpha \cdot \mathbf{B}_s \mathbf{A}_s ,
\end{equation}
for tokens in the prompt that match $s$. Other tokens use the base projection $\mathbf{W}$ without modification.

This strategy ensures that each subject LoRA module influences only its designated tokens, preventing mutual interference and allowing faithful multi-subject rendering. Since LoRA modules are tuned independently but applied jointly, FreeLoRA enables training-free fusion of multiple identities within a single generation.

\begin{table}[t]
    \centering
    \tabcolsep=0.45cm
    \caption{Quantitative comparison of single-subject personalization on the ViCo~\cite{Hao2023ViCo} dataset. Results of other methods are taken from ARBooth~\cite{chung2025fine}.
}
    \begin{tabular}{c c c c c}
    \toprule
    Methods & Type & $\mathbf{I_{dino}}$ & $\mathbf{I_{clip}}$ & $\mathbf{T_{clip}}$ \\
    \midrule
    ELITE~\cite{wei2023elite} & Training-Based & 0.584 & 0.783 & 0.223 \\
    \midrule
    DreamMatcher~\cite{nam2024dreammatcher} & \multirow{7}{*}{Tuning-Based} & 0.682 & 0.823 & 0.234 \\
    Textual Inversion~\cite{gal2022image} & & 0.529 & 0.770 & 0.220 \\
    DreamBooth~\cite{ruiz2023dreambooth} & & 0.640 & 0.815 & 0.236 \\
    Custom Diffusion~\cite{kumari2023multi} & & 0.659 & 0.815 & 0.237 \\
    ViCo~\cite{Hao2023ViCo} & & 0.643 & 0.816 & 0.228 \\
    ARBooth~\cite{chung2025fine} & & 0.705 & 0.824 & \textbf{0.253} \\
    % \midrule
    \textbf{Ours} & & \textbf{0.735} & \textbf{0.837} & 0.249 \\
    \bottomrule
    \end{tabular}
    \label{tab:vico}
\end{table}

\subsection{Discussion}

% FreeLoRA is lightweight and broadly applicable. It requires only minimal architectural modification by injecting LoRA modules into key and value projections of cross-attention layers. This design makes it compatible with transformer-based models across various generation paradigms, including autoregressive and transformer-based diffusion models.

% Despite its simplicity, FreeLoRA enables complex personalization scenarios such as multi-subject generation. We attribute this capability to two core strategies: (1) Full Token Tuning, which encourages the model to learn token-to-image correspondences from limited data, and (2) Subject-Aware Inference, which activates each LoRA only on its associated subject tokens, ensuring localized identity control without affecting unrelated content.

FreeLoRA demonstrates that multi-subject personalization does not require complex model surgery, layout conditioning, or joint optimization across subjects. By carefully selecting where and how to apply lightweight adaptations, such as inserting LoRA into cross-attention layers and activating them based on prompt tokens, our method enables flexible and training-free multi-subject composition.

Interestingly, our method does not impose any explicit constraint to align subject tokens with visual regions. Through the combination of Full Token Tuning (FTT) and Subject-Aware Inference (SAI), the model naturally learns this correspondence. This emergent behavior aligns with Occam’s razor: the network favors the simplest explanation, associating each subject token with its corresponding identity without additional supervision.
\label{exp}
\begin{table}[t]
    \centering
    \tabcolsep=0.45cm
    \caption{Quantitative comparison on the DreamBooth~\cite{ruiz2023dreambooth} dataset for single-subject personalization. Our method outperforms other tuning-based baselines and achieves competitive results compared to training-based methods. Results are taken from UNO~\cite{wu2025less}.}
    \begin{tabular}{c c c c c}
    \toprule
    Methods & Type & $\mathbf{I_{dino}}$ & $\mathbf{I_{clip}}$ & $\mathbf{T_{clip}}$ \\
    \midrule
    ELITE~\cite{wei2023elite} & \multirow{7}{*}{Training-Based} & 0.647 & 0.772 & 0.296 \\
    Re-Imagen~\cite{chen2022re} & & 0.600 & 0.740 & 0.270 \\
    BootPIG~\cite{purushwalkam2024bootpig} & & 0.674 & 0.797 & 0.311 \\
    SSR-Encoder~\cite{zhang2024ssr} & & 0.612 & 0.821 & 0.308 \\
    RealCustom++~\cite{mao2024realcustom++} & & 0.702 & 0.794 & \textbf{0.318} \\
    OmniGen~\cite{xiao2024omnigen} & & 0.693 & 0.801 & \underline{0.315} \\
    OminiControl~\cite{tan2024ominicontrol} & & 0.684 & 0.799 & 0.312 \\
    Subject-Diffusion~\cite{ma2024subject} & & 0.711 & 0.787 & 0.293 \\
    % FLUX.1 IP-Adapter & 0.582 & 0.820 & 0.288 \\
    UNO~\cite{wu2025less} & & \textbf{0.760} & \textbf{0.835} & 0.304 \\
    \midrule
    Textual Inversion~\cite{gal2022image} & \multirow{5}{*}{Tuning-Based} & 0.569 & 0.780 & 0.255 \\
    DreamBooth~\cite{ruiz2023dreambooth} & & 0.668 & 0.803 & 0.305 \\
    BLIP-Diffusion~\cite{li2023blip} & & 0.670 & 0.805 & 0.302 \\
    PersonalAR~\cite{sun2025personalized} & & 0.671 & 0.785 & 0.314 \\
    % \midrule
    \textbf{Ours} & & \underline{0.720} & \underline{0.829} & 0.306 \\
    \bottomrule
    \end{tabular}
    \label{tab:db_single}
\end{table}

\begin{table}[t]
    \centering
    \tabcolsep=0.45cm
    \caption{Quantitative comparison on the DreamBooth~\cite{ruiz2023dreambooth} dataset for multi-subject personalization. Our method outperforms other tuning-based methods and achieves results comparable to the SOTA training-based methods. Results are taken from UNO~\cite{wu2025less}.}
    \begin{tabular}{c c c c c c}
    \toprule
    Methods & Type & $\mathbf{I_{dino}}$ & $\mathbf{I_{clip}}$ & $\mathbf{T_{clip}}$ \\
    \midrule
    Subject Diffusion~\cite{ma2024subject} & \multirow{5}{*}{Training-Based} & 0.506 & 0.696 & 0.310 \\
    MIP-Adapter~\cite{huang2025resolving} & & 0.482 & 0.726 & 0.311 \\
    MS-Diffusion~\cite{wang2024ms} & & 0.525 & 0.726 & 0.319 \\
    OmniGen~\cite{xiao2024omnigen} & & 0.511 & 0.722 &\textbf{0.331} \\
    UNO~\cite{wu2025less} & & \textbf{0.542} & \underline{0.733} & \underline{0.322} \\
    \midrule
    DreamBooth~\cite{ruiz2023dreambooth} & \multirow{3}{*}{Tuning-Based} & 0.430 & 0.695 & 0.308 \\
    % Custom Diffusion & 0.464 & 0.698 & 0.300 \\
    BLIP-Diffusion~\cite{li2023blip} & & 0.464 & 0.698 & 0.300 \\
    % \midrule
    \textbf{Ours} & & \underline{0.529} & \textbf{0.740} & 0.321 \\
    \bottomrule
    \end{tabular}
    \label{tab:db_multi}
\end{table}

\begin{figure}
    \centering
    \includegraphics[width=\linewidth]{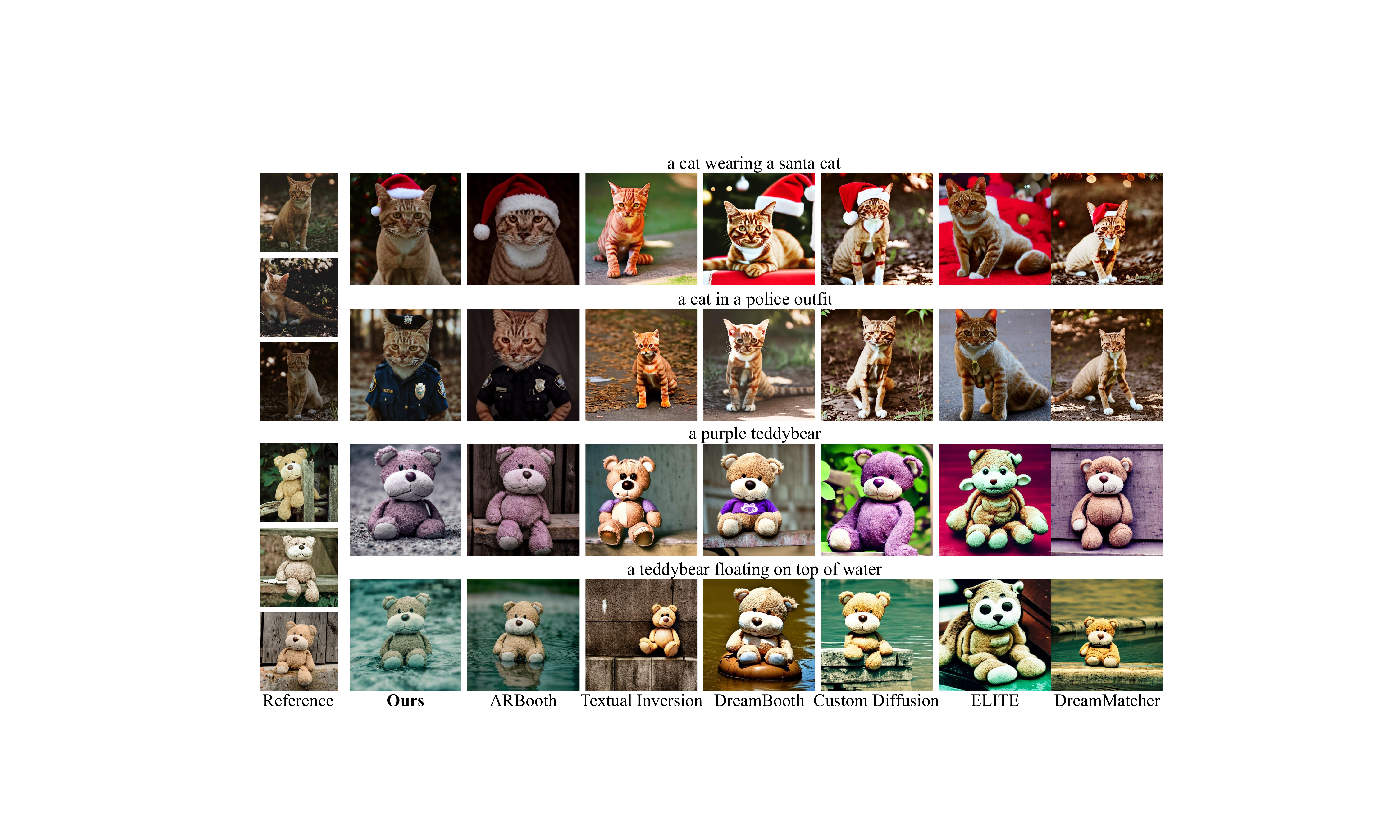}
    \caption{Qualitative comparison of single-subject personalization. Both our method and ARBooth are built on Infinity-2B~\cite{han2024infinity}. Thanks to our Subject-Aware Inference and Full Token Tuning strategies, our method achieves higher visual quality. Results of other methods are taken from ARBooth~\cite{chung2025fine}.}
    \label{fig:single}
\end{figure}

\begin{figure}
    \centering
    \includegraphics[width=\linewidth]{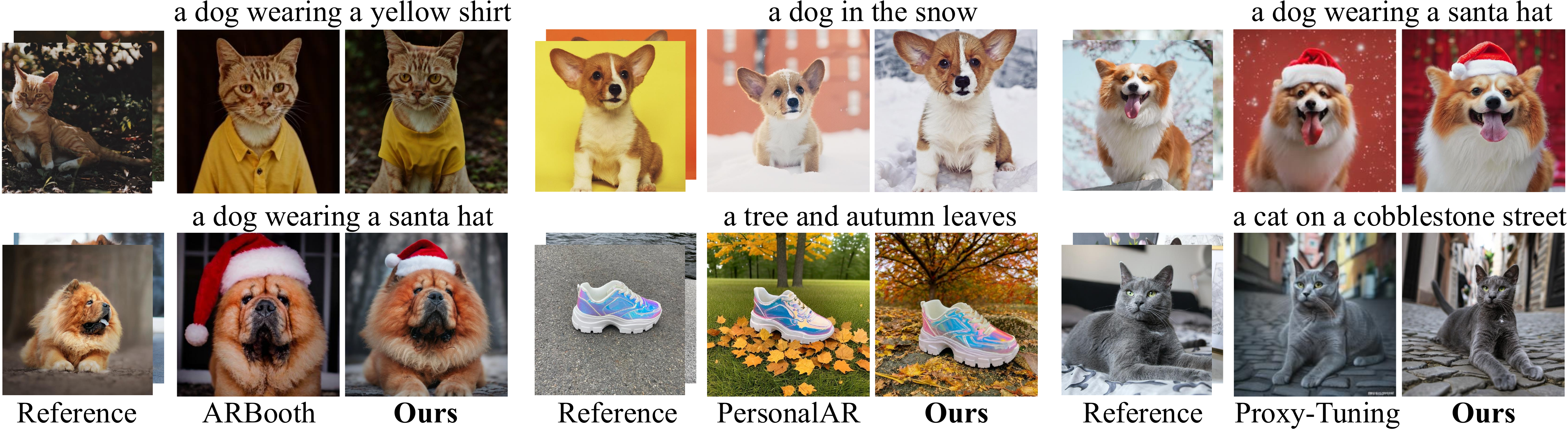}
    \caption{Qualitative comparison of single-subject personalization between autoregressive models. Since the compared AR-based methods are concurrent works with no official code released, their results are directly taken from their papers for reference.}
    \label{fig:single_ar}
\end{figure}

\textbf{Datasets.} Following previous methods~\cite{ruiz2023dreambooth,Hao2023ViCo,wu2025less,chung2025fine}, we conduct our experiments on two widely used datasets: DreamBooth~\cite{ruiz2023dreambooth} and ViCo~\cite{Hao2023ViCo}. The DreamBooth dataset consists of 30 subjects, including 9 living and 21 non-living ones, with approximately 5 images per subject. For single-subject personalization, each subject is paired with 25 prompts. For the multi-subject setting, 30 unique subject pairs are provided, each with 25 prompts. We generate 6 images per prompt to mitigate randomness, resulting in 4,500 images for both tasks.  
The ViCo dataset contains 16 subjects (5 living, 11 non-living), each with 31 prompts. We generate 8 images per prompt, totaling 3,968 images.

\textbf{Evaluation Metrics.} Following prior work~\cite{wu2025less,ruiz2023dreambooth,chung2025fine}, we use both image-to-image and text-to-image similarity metrics to assess subject fidelity and prompt consistency. Specifically, we compute DINO-based~\cite{caron2021emerging} and CLIP-based~\cite{radford2021learning} visual similarities, denoted as $\mathbf{I}_\text{dino}$ and $\mathbf{I}_\text{clip}$, respectively. We also evaluate CLIP-based text-to-image similarity $\mathbf{T}_\text{clip}$ to measure alignment between generated images and input prompts. All metrics are computed using cosine similarity between embeddings and averaged over multiple random seeds and subjects to ensure robustness.

\subsection{Quantitative Comparisons}

\textbf{Single-Subject Personalization.} We compare our method with both training-based and tuning-based baselines on single-subject personalization. Table~\ref{tab:vico} reports results on the ViCo dataset, where our approach surpasses ARBooth~\cite{chung2025fine} in subject fidelity while maintaining comparable prompt consistency. Table~\ref{tab:db_single} shows results on the DreamBooth dataset, where our method achieves SOTA performance among tuning-based methods and performs competitively with the SOTA training-based method UNO~\cite{wu2025less}, which requires costly large-scale training.

\textbf{Multi-Subject Personalization.} Tuning-based methods typically struggle with multi-subject generation due to the need for subject-specific fine-tuning. As shown in Table~\ref{tab:db_multi}, training-based methods generally perform better. However, our proposed FreeLoRA, as a tuning-based approach, achieves results on par with SOTA training-based methods and outperforms all other tuning-based baselines. This improvement is attributed to our Full Token Tuning and Subject-Aware Inference strategies, which are further validated through ablation studies.

\subsection{Qualitative Comparisons}

\begin{figure}
    \centering
    \includegraphics[width=\linewidth]{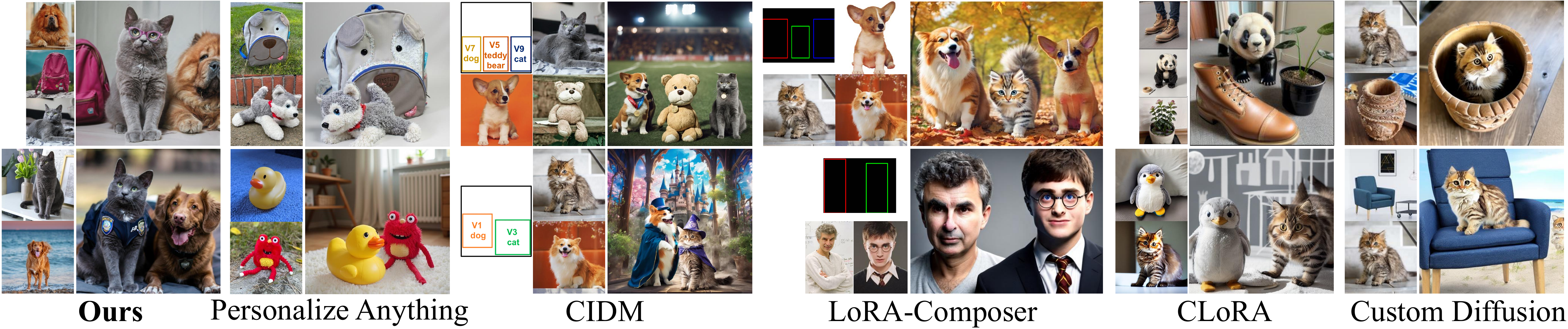}
    \caption{Qualitative comparison of multi-subject personalization among non-training-based methods. Personalize Anything~\cite{feng2025personalize} does not support flexible prompt control. CIDM~\cite{dong2024continually} and LoRA-Composer~\cite{yang2024lora} require layout guidance, while CLoRA~\cite{meral2024clora} and Custom Diffusion~\cite{kumari2023multi} rely on additional optimization. In contrast, our method achieves superior results without requiring layout conditions or further optimization. As non-training-based methods are highly case-sensitive, all results are directly taken from their papers using different prompts to ensure a fair comparison.}
    \label{fig:multi_tuning}
\end{figure}

\begin{figure}
    \centering
    \includegraphics[width=\linewidth]{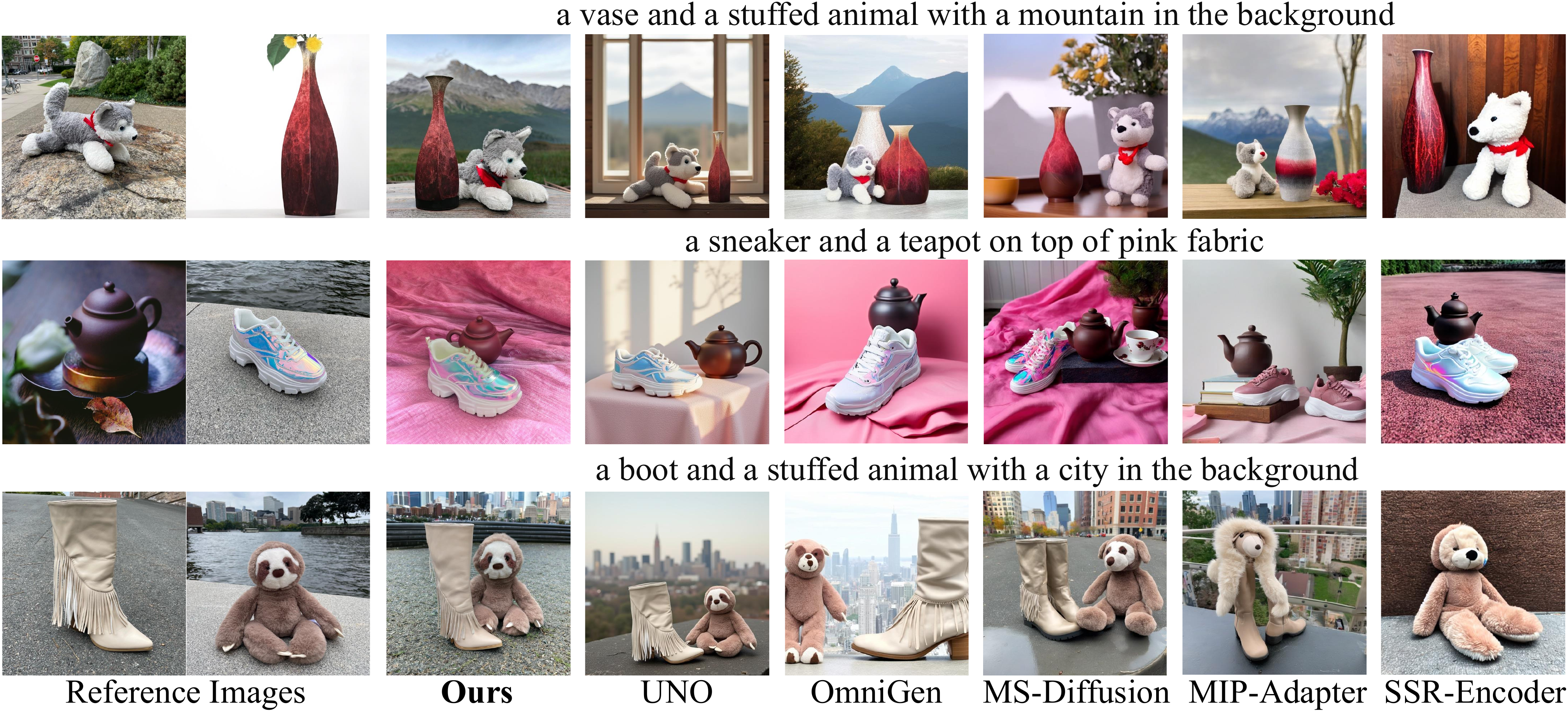}
    \caption{Qualitative comparison of multi-subject personalization with training-based methods. Our method preserves subject details more effectively, as training-based methods depend heavily on data coverage. Results of other methods are taken from UNO~\cite{wu2025less}. Please zoom in for clearer comparison.}
    \label{fig:multi_training}
\end{figure}

\textbf{Single-Subject Personalization.} We conduct qualitative comparisons on single-subject personalization, as shown in Figure~\ref{fig:single}. Both our method and ARBooth are built upon the visual autoregressive model Infinity-2B, and demonstrate superior subject fidelity and prompt consistency compared to previous diffusion-based methods. Furthermore, since our method does not modify the parameters of the pretrained model, it produces higher-quality images than ARBooth. To further demonstrate the superiority of our approach, we compare it with other AR-based methods, as shown in Figure~\ref{fig:single_ar}. Proxy-Tuning utilizes a diffusion-based personalized model to generate proxy images for fine-tuning an AR model, leading to suboptimal image quality. PersonalAR and ARBooth require fine-tuning the pretrained model, whereas our method leverages LoRA to achieve better image quality.

\textbf{Multi-Subject Personalization.} We conduct qualitative comparisons against both training-based and non-training-based methods on multi-subject personalization. Figure~\ref{fig:multi_tuning} shows the results among non-training-based methods, where our approach achieves the best visual quality and subject fidelity. Specifically, the training-free method Personalize Anything~\cite{feng2025personalize} exhibits good visual quality, but is constrained by the pretrained model, leading to reduced subject fidelity and limited prompt flexibility. Other tuning-based methods yield suboptimal results, as they require additional optimization or layout condition to fuse multiple LoRA modules or fine-tuned models. Our method benefits from the Subject-Aware Inference strategy, which enables seamless and training-free fusion of multiple LoRAs. Comparisons with training-based methods are shown in Figure~\ref{fig:multi_training}. Since training-based approaches may suffer from limited generalization, the personalized results may lack subject-specific details. Additionally, they require large-scale training to personalize different types of subjects.

\subsection{Ablation Studies}

\begin{figure}
    \centering
    \includegraphics[width=\linewidth]{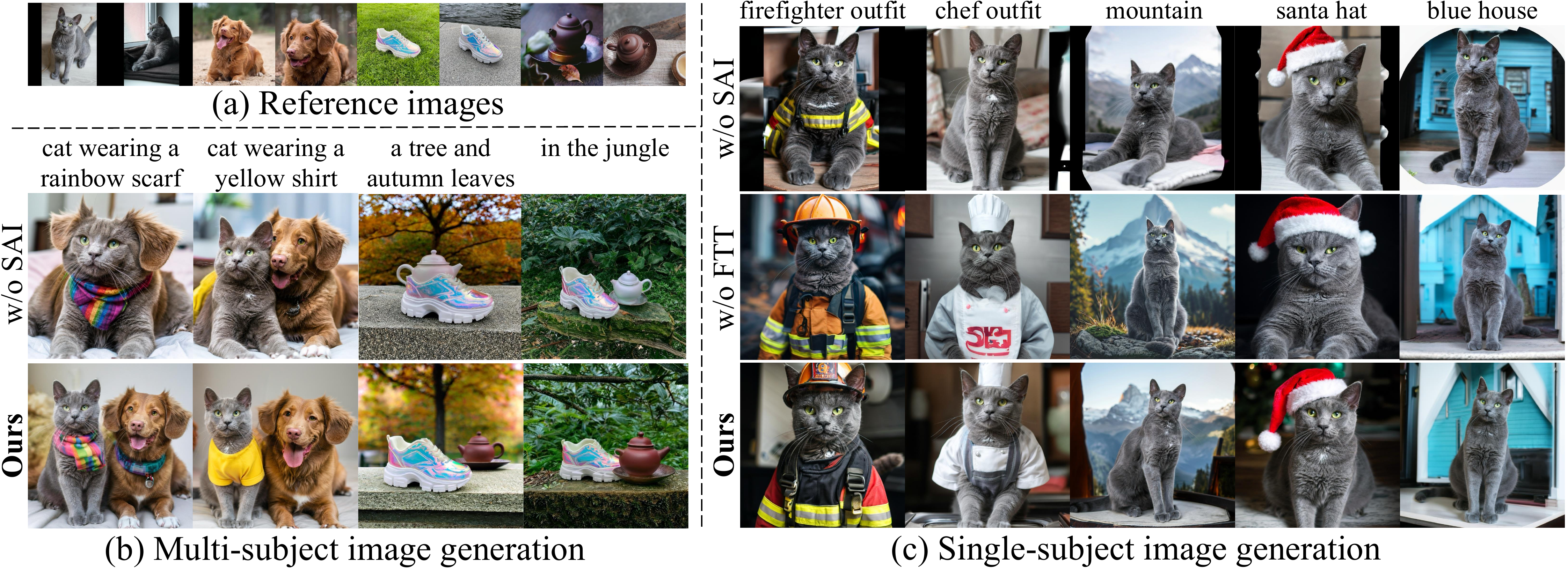}
    \caption{Qualitative ablation study on Subject-Aware Inference (SAI) and Full Token Tuning (FTT). Removing SAI leads to background overfitting and mutual interference between LoRA modules in multi-subject scenarios. FTT improves visual fidelity by encouraging token-to-content alignment.}
    \label{fig:ablation}
\end{figure}

\textbf{Effect of Subject-Aware Inference.}  
To assess the impact of our proposed Subject-Aware Inference (SAI) strategy, we conduct quantitative ablation studies as shown in Table~\ref{tab:ablation}. The results indicate that removing SAI leads to higher visual similarity with reference images, but at the expense of reduced prompt consistency. We attribute this increased similarity to overfitting on the reference images, particularly their backgrounds. This hypothesis is further supported by the qualitative results in Figure~\ref{fig:ablation}(c), where the generated images without SAI clearly exhibit background overfitting. Moreover, the results in Figure~\ref{fig:ablation}(b) demonstrate the necessity of SAI in multi-subject personalization scenarios, as it helps avoid mutual interference between different LoRA modules.

\textbf{Effect of Full Token Tuning.}  
In the previous ablation study, we validated the effectiveness of Subject-Aware Inference. What if we apply the same strategy during tuning? The quantitative results in Table~\ref{tab:ablation}, along with the qualitative examples in Figure~\ref{fig:ablation}, underscore the importance of Full Token Tuning strategy. Specifically, tuning solely on subject tokens forces them to encode the entire image content including background and context rather than just subject-specific features, which significantly degrades image quality. In contrast, tuning on all tokens enables the LoRA modules to automatically learn the mapping between image content and corresponding tokens, allowing for more accurate subject representation and improved visual fidelity.

\begin{table}[t]
    \centering
    \tabcolsep=0.36cm
    \caption{Quantitative ablation study on Subject-Aware Inference (SAI) and Full Token Tuning (FTT). We evaluate single-subject personalization on the ViCo~\cite{Hao2023ViCo} dataset and multi-subject personalization on the DreamBooth~\cite{ruiz2023dreambooth} dataset. Removing SAI increases image similarity but reduces prompt consistency due to overfitting. Removing FTT degrades subject fidelity.}
    \begin{tabular}{ccccccc}
    \toprule
    \multirow{2}{*}{Methods} & \multicolumn{3}{c}{Single-Subject} & \multicolumn{3}{c}{Multi-Subject} \\
    \cmidrule(lr){2-4} \cmidrule(lr){5-7}
                             & $\mathbf{I_{dino}}$ & $\mathbf{I_{clip}}$ & $\mathbf{T_{clip}}$ & $\mathbf{I_{dino}}$ & $\mathbf{I_{clip}}$ & $\mathbf{T_{clip}}$ \\
    \midrule
    w/o Subject-Aware Inference & \textbf{0.744} & \textbf{0.846} & 0.246 & \textbf{0.543} & \textbf{0.748} & 0.308 \\
    w/o Full Token Tuning       & 0.673 & 0.817 & \textbf{0.256} & 0.502 & 0.726 & \textbf{0.328} \\
    \textbf{Ours}                        & \underline{0.735} & \underline{0.837} & \underline{0.249} & \underline{0.529} & \underline{0.740} & \underline{0.321} \\
    \bottomrule
    \end{tabular}
    \label{tab:ablation}
\end{table}

\section{Conclusion}
\label{conclusion}
We propose FreeLoRA, a simple yet effective framework for multi-subject personalization in autoregressive image generation. FreeLoRA adapts subject-specific LoRA modules using a Full Token Tuning (FTT) strategy and applies them at inference via Subject-Aware Inference (SAI), selectively activating each module based on its corresponding subject tokens. This design enables training-free fusion of multiple personalized subjects within a single prompt, while mitigating overfitting and inter-subject interference. 
% Our method requires no modification to the pretrained backbone and can be seamlessly integrated into Transformer-based architectures.
Extensive experiments on standard benchmarks demonstrate the strong generalization and practical utility of FreeLoRA across both single-subject and multi-subject personalization tasks. 
% Extensive experiments demonstrate the strong generalization and practical utility of FreeLoRA across both single-subject and multi-subject personalization tasks. 
Nevertheless, as our method relies on adapting a pretrained model, it may fail in complex multi-subject scenes where the pretrained model lacks sufficient compositional capability, which further highlights the importance of advancing multimodal foundation models.

\bibliographystyle{plain}
\bibliography{ref}

\begin{thebibliography}{10}

\bibitem{caron2021emerging}
Mathilde Caron, Hugo Touvron, Ishan Misra, Herv{\'e} J{\'e}gou, Julien Mairal, Piotr Bojanowski, and Armand Joulin.
\newblock Emerging properties in self-supervised vision transformers.
\newblock In {\em Proceedings of the IEEE/CVF international conference on computer vision}, pages 9650--9660, 2021.

\bibitem{chen2022re}
Wenhu Chen, Hexiang Hu, Chitwan Saharia, and William~W Cohen.
\newblock Re-imagen: Retrieval-augmented text-to-image generator.
\newblock {\em arXiv preprint arXiv:2209.14491}, 2022.

\bibitem{chung2025fine}
Jiwoo Chung, Sangeek Hyun, Hyunjun Kim, Eunseo Koh, MinKyu Lee, and Jae-Pil Heo.
\newblock Fine-tuning visual autoregressive models for subject-driven generation.
\newblock {\em arXiv preprint arXiv:2504.02612}, 2025.

\bibitem{dhariwal2021diffusion}
Prafulla Dhariwal and Alexander Nichol.
\newblock Diffusion models beat gans on image synthesis.
\newblock {\em Advances in neural information processing systems}, 34:8780--8794, 2021.

\bibitem{ding2024freecustom}
Ganggui Ding, Canyu Zhao, Wen Wang, Zhen Yang, Zide Liu, Hao Chen, and Chunhua Shen.
\newblock Freecustom: Tuning-free customized image generation for multi-concept composition.
\newblock In {\em Proceedings of the IEEE/CVF Conference on Computer Vision and Pattern Recognition}, pages 9089--9098, 2024.

\bibitem{dong2024continually}
Jiahua Dong, Wenqi Liang, Hongliu Li, Duzhen Zhang, Meng Cao, Henghui Ding, Salman~H Khan, and Fahad Shahbaz~Khan.
\newblock How to continually adapt text-to-image diffusion models for flexible customization?
\newblock {\em Advances in Neural Information Processing Systems}, 37:130057--130083, 2024.

\bibitem{feng2025personalize}
Haoran Feng, Zehuan Huang, Lin Li, Hairong Lv, and Lu~Sheng.
\newblock Personalize anything for free with diffusion transformer.
\newblock {\em arXiv preprint arXiv:2503.12590}, 2025.

\bibitem{gal2022image}
Rinon Gal, Yuval Alaluf, Yuval Atzmon, Or~Patashnik, Amit~H Bermano, Gal Chechik, and Daniel Cohen-Or.
\newblock An image is worth one word: Personalizing text-to-image generation using textual inversion.
\newblock {\em arXiv preprint arXiv:2208.01618}, 2022.

\bibitem{gu2023mix}
Yuchao Gu, Xintao Wang, Jay~Zhangjie Wu, Yujun Shi, Yunpeng Chen, Zihan Fan, Wuyou Xiao, Rui Zhao, Shuning Chang, Weijia Wu, et~al.
\newblock Mix-of-show: Decentralized low-rank adaptation for multi-concept customization of diffusion models.
\newblock {\em Advances in Neural Information Processing Systems}, 36:15890--15902, 2023.

\bibitem{han2024infinity}
Jian Han, Jinlai Liu, Yi~Jiang, Bin Yan, Yuqi Zhang, Zehuan Yuan, Bingyue Peng, and Xiaobing Liu.
\newblock Infinity: Scaling bitwise autoregressive modeling for high-resolution image synthesis.
\newblock {\em arXiv preprint arXiv:2412.04431}, 2024.

\bibitem{Hao2023ViCo}
Shaozhe Hao, Kai Han, Shihao Zhao, and Kwan-Yee~K. Wong.
\newblock Vico: Detail-preserving visual condition for personalized text-to-image generation.
\newblock 2023.

\bibitem{he2025anystory}
Junjie He, Yuxiang Tuo, Binghui Chen, Chongyang Zhong, Yifeng Geng, and Liefeng Bo.
\newblock Anystory: Towards unified single and multiple subject personalization in text-to-image generation.
\newblock {\em arXiv preprint arXiv:2501.09503}, 2025.

\bibitem{hu2022lora}
Edward~J Hu, Yelong Shen, Phillip Wallis, Zeyuan Allen-Zhu, Yuanzhi Li, Shean Wang, Lu~Wang, Weizhu Chen, et~al.
\newblock Lora: Low-rank adaptation of large language models.
\newblock {\em ICLR}, 1(2):3, 2022.

\bibitem{huang2025resolving}
Qihan Huang, Siming Fu, Jinlong Liu, Hao Jiang, Yipeng Yu, and Jie Song.
\newblock Resolving multi-condition confusion for finetuning-free personalized image generation.
\newblock In {\em Proceedings of the AAAI Conference on Artificial Intelligence}, volume~39, pages 3707--3714, 2025.

\bibitem{kang2025flux}
Hao Kang, Stathi Fotiadis, Liming Jiang, Qing Yan, Yumin Jia, Zichuan Liu, Min~Jin Chong, and Xin Lu.
\newblock Flux already knows-activating subject-driven image generation without training.
\newblock {\em arXiv preprint arXiv:2504.11478}, 2025.

\bibitem{kumari2023multi}
Nupur Kumari, Bingliang Zhang, Richard Zhang, Eli Shechtman, and Jun-Yan Zhu.
\newblock Multi-concept customization of text-to-image diffusion.
\newblock In {\em Proceedings of the IEEE/CVF conference on computer vision and pattern recognition}, pages 1931--1941, 2023.

\bibitem{li2023blip}
Dongxu Li, Junnan Li, and Steven Hoi.
\newblock Blip-diffusion: Pre-trained subject representation for controllable text-to-image generation and editing.
\newblock {\em Advances in Neural Information Processing Systems}, 36:30146--30166, 2023.

\bibitem{li2025autoregressive}
Tianhong Li, Yonglong Tian, He~Li, Mingyang Deng, and Kaiming He.
\newblock Autoregressive image generation without vector quantization.
\newblock {\em Advances in Neural Information Processing Systems}, 37:56424--56445, 2025.

\bibitem{liu2023cones}
Zhiheng Liu, Ruili Feng, Kai Zhu, Yifei Zhang, Kecheng Zheng, Yu~Liu, Deli Zhao, Jingren Zhou, and Yang Cao.
\newblock Cones: Concept neurons in diffusion models for customized generation.
\newblock {\em arXiv preprint arXiv:2303.05125}, 2023.

\bibitem{ma2024subject}
Jian Ma, Junhao Liang, Chen Chen, and Haonan Lu.
\newblock Subject-diffusion: Open domain personalized text-to-image generation without test-time fine-tuning.
\newblock In {\em ACM SIGGRAPH 2024 Conference Papers}, pages 1--12, 2024.

\bibitem{mao2024realcustom++}
Zhendong Mao, Mengqi Huang, Fei Ding, Mingcong Liu, Qian He, and Yongdong Zhang.
\newblock Realcustom++: Representing images as real-word for real-time customization.
\newblock {\em arXiv preprint arXiv:2408.09744}, 2024.

\bibitem{meral2024clora}
Tuna Han~Salih Meral, Enis Simsar, Federico Tombari, and Pinar Yanardag.
\newblock Clora: A contrastive approach to compose multiple lora models.
\newblock {\em arXiv preprint arXiv:2403.19776}, 2024.

\bibitem{mou2025dreamo}
Chong Mou, Yanze Wu, Wenxu Wu, Zinan Guo, Pengze Zhang, Yufeng Cheng, Yiming Luo, Fei Ding, Shiwen Zhang, Xinghui Li, et~al.
\newblock Dreamo: A unified framework for image customization.
\newblock {\em arXiv preprint arXiv:2504.16915}, 2025.

\bibitem{nam2024dreammatcher}
Jisu Nam, Heesu Kim, DongJae Lee, Siyoon Jin, Seungryong Kim, and Seunggyu Chang.
\newblock Dreammatcher: appearance matching self-attention for semantically-consistent text-to-image personalization.
\newblock In {\em Proceedings of the IEEE/CVF Conference on Computer Vision and Pattern Recognition}, pages 8100--8110, 2024.

\bibitem{ouyang2025k}
Ziheng Ouyang, Zhen Li, and Qibin Hou.
\newblock K-lora: Unlocking training-free fusion of any subject and style loras.
\newblock {\em arXiv preprint arXiv:2502.18461}, 2025.

\bibitem{podell2023sdxl}
Dustin Podell, Zion English, Kyle Lacey, Andreas Blattmann, Tim Dockhorn, Jonas M{\"u}ller, Joe Penna, and Robin Rombach.
\newblock Sdxl: Improving latent diffusion models for high-resolution image synthesis.
\newblock {\em arXiv preprint arXiv:2307.01952}, 2023.

\bibitem{purushwalkam2024bootpig}
Senthil Purushwalkam, Akash Gokul, Shafiq Joty, and Nikhil Naik.
\newblock Bootpig: Bootstrapping zero-shot personalized image generation capabilities in pretrained diffusion models.
\newblock {\em arXiv preprint arXiv:2401.13974}, 2024.

\bibitem{radford2021learning}
Alec Radford, Jong~Wook Kim, Chris Hallacy, Aditya Ramesh, Gabriel Goh, Sandhini Agarwal, Girish Sastry, Amanda Askell, Pamela Mishkin, Jack Clark, et~al.
\newblock Learning transferable visual models from natural language supervision.
\newblock In {\em International conference on machine learning}, pages 8748--8763. PmLR, 2021.

\bibitem{rombach2022high}
Robin Rombach, Andreas Blattmann, Dominik Lorenz, Patrick Esser, and Bj{\"o}rn Ommer.
\newblock High-resolution image synthesis with latent diffusion models.
\newblock In {\em Proceedings of the IEEE/CVF conference on computer vision and pattern recognition}, pages 10684--10695, 2022.

\bibitem{ruiz2023dreambooth}
Nataniel Ruiz, Yuanzhen Li, Varun Jampani, Yael Pritch, Michael Rubinstein, and Kfir Aberman.
\newblock Dreambooth: Fine tuning text-to-image diffusion models for subject-driven generation.
\newblock In {\em Proceedings of the IEEE/CVF conference on computer vision and pattern recognition}, pages 22500--22510, 2023.

\bibitem{lora_stable}
Simo Ryu.
\newblock Low-rank adaptation for fast text-to-image diffusion fine-tuning.
\newblock \url{https://github.com/cloneofsimo/lora}.

\bibitem{shah2024ziplora}
Viraj Shah, Nataniel Ruiz, Forrester Cole, Erika Lu, Svetlana Lazebnik, Yuanzhen Li, and Varun Jampani.
\newblock Ziplora: Any subject in any style by effectively merging loras.
\newblock In {\em European Conference on Computer Vision}, pages 422--438. Springer, 2024.

\bibitem{shi2024instantbooth}
Jing Shi, Wei Xiong, Zhe Lin, and Hyun~Joon Jung.
\newblock Instantbooth: Personalized text-to-image generation without test-time finetuning.
\newblock In {\em Proceedings of the IEEE/CVF conference on computer vision and pattern recognition}, pages 8543--8552, 2024.

\bibitem{shi2024llamafusion}
Weijia Shi, Xiaochuang Han, Chunting Zhou, Weixin Liang, Xi~Victoria Lin, Luke Zettlemoyer, and Lili Yu.
\newblock Llamafusion: Adapting pretrained language models for multimodal generation.
\newblock {\em arXiv preprint arXiv:2412.15188}, 2024.

\bibitem{shin2024large}
Chaehun Shin, Jooyoung Choi, Heeseung Kim, and Sungroh Yoon.
\newblock Large-scale text-to-image model with inpainting is a zero-shot subject-driven image generator.
\newblock {\em arXiv preprint arXiv:2411.15466}, 2024.

\bibitem{sun2025personalized}
Kaiyue Sun, Xian Liu, Yao Teng, and Xihui Liu.
\newblock Personalized text-to-image generation with auto-regressive models.
\newblock {\em arXiv preprint arXiv:2504.13162}, 2025.

\bibitem{tan2024ominicontrol}
Zhenxiong Tan, Songhua Liu, Xingyi Yang, Qiaochu Xue, and Xinchao Wang.
\newblock Ominicontrol: Minimal and universal control for diffusion transformer.
\newblock {\em arXiv preprint arXiv:2411.15098}, 2024.

\bibitem{team2024chameleon}
Chameleon Team.
\newblock Chameleon: Mixed-modal early-fusion foundation models.
\newblock {\em arXiv preprint arXiv:2405.09818}, 2024.

\bibitem{team2023gemini}
Gemini Team, Rohan Anil, Sebastian Borgeaud, Jean-Baptiste Alayrac, Jiahui Yu, Radu Soricut, Johan Schalkwyk, Andrew~M Dai, Anja Hauth, Katie Millican, et~al.
\newblock Gemini: a family of highly capable multimodal models.
\newblock {\em arXiv preprint arXiv:2312.11805}, 2023.

\bibitem{tian2025visual}
Keyu Tian, Yi~Jiang, Zehuan Yuan, Bingyue Peng, and Liwei Wang.
\newblock Visual autoregressive modeling: Scalable image generation via next-scale prediction.
\newblock {\em Advances in neural information processing systems}, 37:84839--84865, 2025.

\bibitem{tong2024metamorph}
Shengbang Tong, David Fan, Jiachen Zhu, Yunyang Xiong, Xinlei Chen, Koustuv Sinha, Michael Rabbat, Yann LeCun, Saining Xie, and Zhuang Liu.
\newblock Metamorph: Multimodal understanding and generation via instruction tuning.
\newblock {\em arXiv preprint arXiv:2412.14164}, 2024.

\bibitem{wang2024ms}
Xierui Wang, Siming Fu, Qihan Huang, Wanggui He, and Hao Jiang.
\newblock Ms-diffusion: Multi-subject zero-shot image personalization with layout guidance.
\newblock {\em arXiv preprint arXiv:2406.07209}, 2024.

\bibitem{wang2024emu3}
Xinlong Wang, Xiaosong Zhang, Zhengxiong Luo, Quan Sun, Yufeng Cui, Jinsheng Wang, Fan Zhang, Yueze Wang, Zhen Li, Qiying Yu, et~al.
\newblock Emu3: Next-token prediction is all you need.
\newblock {\em arXiv preprint arXiv:2409.18869}, 2024.

\bibitem{weber2024maskbit}
Mark Weber, Lijun Yu, Qihang Yu, Xueqing Deng, Xiaohui Shen, Daniel Cremers, and Liang-Chieh Chen.
\newblock Maskbit: Embedding-free image generation via bit tokens.
\newblock {\em arXiv preprint arXiv:2409.16211}, 2024.

\bibitem{wei2023elite}
Yuxiang Wei, Yabo Zhang, Zhilong Ji, Jinfeng Bai, Lei Zhang, and Wangmeng Zuo.
\newblock Elite: Encoding visual concepts into textual embeddings for customized text-to-image generation.
\newblock In {\em Proceedings of the IEEE/CVF International Conference on Computer Vision}, pages 15943--15953, 2023.

\bibitem{wu2025less}
Shaojin Wu, Mengqi Huang, Wenxu Wu, Yufeng Cheng, Fei Ding, and Qian He.
\newblock Less-to-more generalization: Unlocking more controllability by in-context generation.
\newblock {\em arXiv preprint arXiv:2504.02160}, 2025.

\bibitem{wu2024vila}
Yecheng Wu, Zhuoyang Zhang, Junyu Chen, Haotian Tang, Dacheng Li, Yunhao Fang, Ligeng Zhu, Enze Xie, Hongxu Yin, Li~Yi, et~al.
\newblock Vila-u: a unified foundation model integrating visual understanding and generation.
\newblock {\em arXiv preprint arXiv:2409.04429}, 2024.

\bibitem{wu2025proxy}
Yi~Wu, Lingting Zhu, Lei Liu, Wandi Qiao, Ziqiang Li, Lequan Yu, and Bin Li.
\newblock Proxy-tuning: Tailoring multimodal autoregressive models for subject-driven image generation.
\newblock {\em arXiv preprint arXiv:2503.10125}, 2025.

\bibitem{xiao2024omnigen}
Shitao Xiao, Yueze Wang, Junjie Zhou, Huaying Yuan, Xingrun Xing, Ruiran Yan, Chaofan Li, Shuting Wang, Tiejun Huang, and Zheng Liu.
\newblock Omnigen: Unified image generation.
\newblock {\em arXiv preprint arXiv:2409.11340}, 2024.

\bibitem{xie2024show}
Jinheng Xie, Weijia Mao, Zechen Bai, David~Junhao Zhang, Weihao Wang, Kevin~Qinghong Lin, Yuchao Gu, Zhijie Chen, Zhenheng Yang, and Mike~Zheng Shou.
\newblock Show-o: One single transformer to unify multimodal understanding and generation.
\newblock {\em arXiv preprint arXiv:2408.12528}, 2024.

\bibitem{yang2024lora}
Yang Yang, Wen Wang, Liang Peng, Chaotian Song, Yao Chen, Hengjia Li, Xiaolong Yang, Qinglin Lu, Deng Cai, Boxi Wu, et~al.
\newblock Lora-composer: Leveraging low-rank adaptation for multi-concept customization in training-free diffusion models.
\newblock {\em arXiv preprint arXiv:2403.11627}, 2024.

\bibitem{yu2024randomized}
Qihang Yu, Ju~He, Xueqing Deng, Xiaohui Shen, and Liang-Chieh Chen.
\newblock Randomized autoregressive visual generation.
\newblock {\em arXiv preprint arXiv:2411.00776}, 2024.

\bibitem{zhang2025bringing}
Yuxin Zhang, Minyan Luo, Weiming Dong, Xiao Yang, Haibin Huang, Chongyang Ma, Oliver Deussen, Tong-Yee Lee, and Changsheng Xu.
\newblock Bringing characters to new stories: Training-free theme-specific image generation via dynamic visual prompting.
\newblock {\em arXiv preprint arXiv:2501.15641}, 2025.

\bibitem{zhang2024ssr}
Yuxuan Zhang, Yiren Song, Jiaming Liu, Rui Wang, Jinpeng Yu, Hao Tang, Huaxia Li, Xu~Tang, Yao Hu, Han Pan, et~al.
\newblock Ssr-encoder: Encoding selective subject representation for subject-driven generation.
\newblock In {\em Proceedings of the IEEE/CVF Conference on Computer Vision and Pattern Recognition}, pages 8069--8078, 2024.

\bibitem{zhong2024multi}
Ming Zhong, Yelong Shen, Shuohang Wang, Yadong Lu, Yizhu Jiao, Siru Ouyang, Donghan Yu, Jiawei Han, and Weizhu Chen.
\newblock Multi-lora composition for image generation.
\newblock {\em arXiv preprint arXiv:2402.16843}, 2024.

\bibitem{zhou2024transfusion}
Chunting Zhou, Lili Yu, Arun Babu, Kushal Tirumala, Michihiro Yasunaga, Leonid Shamis, Jacob Kahn, Xuezhe Ma, Luke Zettlemoyer, and Omer Levy.
\newblock Transfusion: Predict the next token and diffuse images with one multi-modal model.
\newblock {\em arXiv preprint arXiv:2408.11039}, 2024.

\bibitem{zhou2024magictailor}
Donghao Zhou, Jiancheng Huang, Jinbin Bai, Jiaze Wang, Hao Chen, Guangyong Chen, Xiaowei Hu, and Pheng-Ann Heng.
\newblock Magictailor: Component-controllable personalization in text-to-image diffusion models.
\newblock {\em arXiv preprint arXiv:2410.13370}, 2024.

\end{thebibliography}

\end{document}